\documentclass[a4paper]{article}

\usepackage{INTERSPEECH2021}
\usepackage{dirtree}
\usepackage{indentfirst}
\usepackage{tabularx}
\usepackage{threeparttable}
\usepackage[hidelinks]{hyperref}

\title{Augmenting End-to-End Steering Angle Prediction with CAN Bus Data}
\name{Amit Singh}

\email{}
\begin{document}
\maketitle  
\begin{abstract}
In recent years, end-to-end steering prediction for autonomous vehicles has become a major area of research. The primary method for achieving end-to-end steering was to use computer vision models on a live feed of video data. However, to further increase accuracy, many companies have added data from light detection and ranging (LiDAR) and/or radar sensors through sensor fusion. However, the addition of lasers and sensors comes at a high financial cost. 

In this paper, I address both of these issues by increasing the accuracy of the computer vision models without the increased cost of using LiDAR and/or sensors. I achieved this by improving the accuracy of computer vision models by sensor fusing CAN bus data, a vehicle protocol, with video data. CAN bus data is a rich source of information about the vehicle’s state, including its speed, steering angle, and acceleration. By fusing this data with video data, the accuracy of the computer vision model’s predictions can be improved. 

The data was collected with a dashboard mounted camera as well as a computer connected to the CAN Bus Wires. I then trained NVIDIA’s DAVE 2 model on the video data and CAN bus data and compared this model to one trained without CAN bus data. The metric I chose to measure success in the experiment is the Root Mean Square Error (RMSE). When I trained the model without CAN bus data, I obtained an RMSE of 0.02492, while the model trained with the CAN bus data achieved an RMSE of 0.01970. This finding indicates that fusing CAN Bus data with video data can reduce the computer vision model’s prediction error by 20\%. 

Furthermore, additional models were tested to determine if they also had a decrease in error when fused with CAN Bus data. I found that large computer vision models performed better with the addition of CAN Bus data. For example, RESNET50’s error decreased by 52\%, EfficientNetB7’s error decreased by 70\%, VGG19’s error decreased by 25\%, NasNetLarge’s error decreased by 80\%, and RESENT152V2’s error decreased by 70\%. These results suggest that fusing CAN Bus data with video data can reduce the error rate of computer vision models. 

Since the results are better with CAN bus data, it shows that CAN bus data can improve accuracy while lowering financial costs. This is because CAN bus data is a relatively inexpensive source of data, and it can be collected using off the-shelf hardware. I believe that training end-to-end steering prediction models with both CAN Bus data and video data has the potential to improve the safety and performance of autonomous vehicles.

\end{abstract}


\section{Introduction}
The idea of self-driving cars was first conceived in the late 1930s, but it was not until the 1980s that they began to be developed at Carnegie Mellon University. When neural networks were first introduced into autonomous vehicles, the main source of data was from cameras. Over time, computers became more powerful and neural networks became larger. With more computational power, autonomous driving models were able to handle other sources of data, such as Radio Detection and Ranging (RADAR) and Light Detection and Ranging (LiDAR). The addition of sensors improved accuracy and safety; however, the sensors also increased the cost of autonomous vehicles, which led many companies to balance cost and accuracy. 

Companies like Tesla aim to make self-driving cars highly safe and more affordable, as evidenced by the highly marketed Tesla Model 3 and the upcoming Tesla Hatchback. In contrast, companies like Google value accuracy over price and continue to outfit their cars with the most up-to-date and expensive sensors available. The aim of this paper is to research methods that allow a balance of cost and accuracy. Controller Area Network (CAN) bus data can be added to a typical computer vision model to decrease error rate while maintaining costs. 

As cars became more popular, regulations related to the safety of cars also increased. One such regulation is the On-Board Diagnostic II (OBDII) standard, which introduced the CAN bus protocol. CAN bus allows a car’s Electronic Control Units (ECUs), which are connected to the different sensors in the car, to communicate without complex wiring through the use of high and low voltage signals. Due to the OBDII regulations, all gas-powered vehicles are required to have an OBDII port and the majority of electric vehicles also have OBDII ports. This allows external hardware to  be connected to the car in order to collect diagnostic packets of information from the CAN bus wires. As a result, the process of collecting data from a car’s onboard sensors is fairly simple. The data from these sensors can then be added to the end-to-end steering angle prediction model to increase its accuracy. 

While sensor fusion techniques exist to combine computer vision models with other types of data, no widely used technique uses CAN Bus data. The contribution this paper makes is by adding CAN bus data to machine learning models to decrease error. 

\section{Related Work}
\label{sec:related}
\begin{figure}[h]
  \includegraphics[width=\columnwidth]{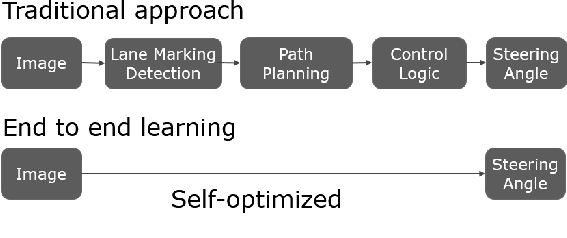}
  \caption{end-to-end break down \cite{7995975}}
  \label{fig:endtoend}
\end{figure}

The research in the steering angle prediction has been divided. Researchers with access to LIDAR and RADAR are proponents of them and use them for regular steering prediction. However, researchers without access to expensive sensors are proponents of end-to-end steering prediction \cite{https://doi.org/10.48550/arxiv.1504.01716}. The difference between these steering prediction models, regular steering prediction and end-to-end steering prediction, is that in the latter, a model is trained to predict the steering angle, while in the former, many smaller models are trained on distinct parts of the data like the weather conditions or the road lanes. This is can be shown in \autoref{fig:endtoend}.

NVIDIA made a significant contribution to the field of end-to-end steering prediction with publication of the DAVE-2 model~\cite{https://doi.org/10.48550/arxiv.1604.07316}. DAVE-2 is a relatively simple Convolution Neural Network (CNN) model that predicts the steering angle of a vehicle based on three camera angles: front, left, and right. NVIDIA was able to achieve low error with the model, and they also released the Udacity dataset that they trained their model on, which has allowed other researchers to optimize and develop more advanced models~\cite{https://doi.org/10.48550/arxiv.1704.07911,Sallab_2017}.
\begin{table*}[h]
  \centering
  \renewcommand{\arraystretch}{1.2}
   
  \begin{tabular}{|c|c|c|c|c|}
    \hline
\textbf{Voltage} & \textbf{Current} & \textbf{Power} & \textbf{Steering Speed} & \textbf{Speed} \\
\hline
    400.368000&-110.420000&-44.208831&2.583333&35.790981\\\hline
    400.584000&-112.580000&-45.097740&0.000000&35.592142\\\hline
    400.810000&-113.600000&-45.532016&2.750000&35.492723\\ \hline
    401.810000&-119.300000&-47.935933&6.750000&34.647658\\ \hline
    401.620000&-118.950000&-47.772712&8.500000&34.597948\\ \hline
    401.847500&-119.125000&-47.870120&9.250000&34.548238\\ \hline
  \end{tabular}
  \caption{Sample Preprocessed CAN Bus Data}
  \label{tab:1}
\end{table*}

Many researchers have written papers which try to calculate the model which predicts the best steering angle. A paper compared four different models to determine which model predicts the best end to end steering angle: Predict 0, 3D LSTM, RESNET 50, and NVIDIA’s DAVE-2 model~\cite{https://doi.org/10.48550/arxiv.1912.05440}. It found that RESNET 50 had the lowest loss. However, creating and determining which model provides the best accuracy is very labor intensive. This paper also used an alternate method to decrease loss, this paper proposes to add CAN bus data to the model which results in the greatest decrease in loss. 

Though there are no other end-to-end steering papers that sensor fuse on both CAN Bus data and image data, there are other papers that sensor fuse image data with LIDAR and or RADAR data~\cite{9011341}. This paper was able to decrease the error by fusing LIDAR and image data together when compared to model training on only image data or only LIDAR data.

In recent years, there has been a development in implementing steering angle prediction models with other models like course planning. Though this paper only goes into the steering angle part, a paper called DiffStack is able to combine modules like steering angle prediction and course planning while still training them under the same neural network \cite{karkus2022diffstack}.

\section{Methodology}

\subsection{Dataset}
\begin{figure}[h]
  \includegraphics[width=\columnwidth]{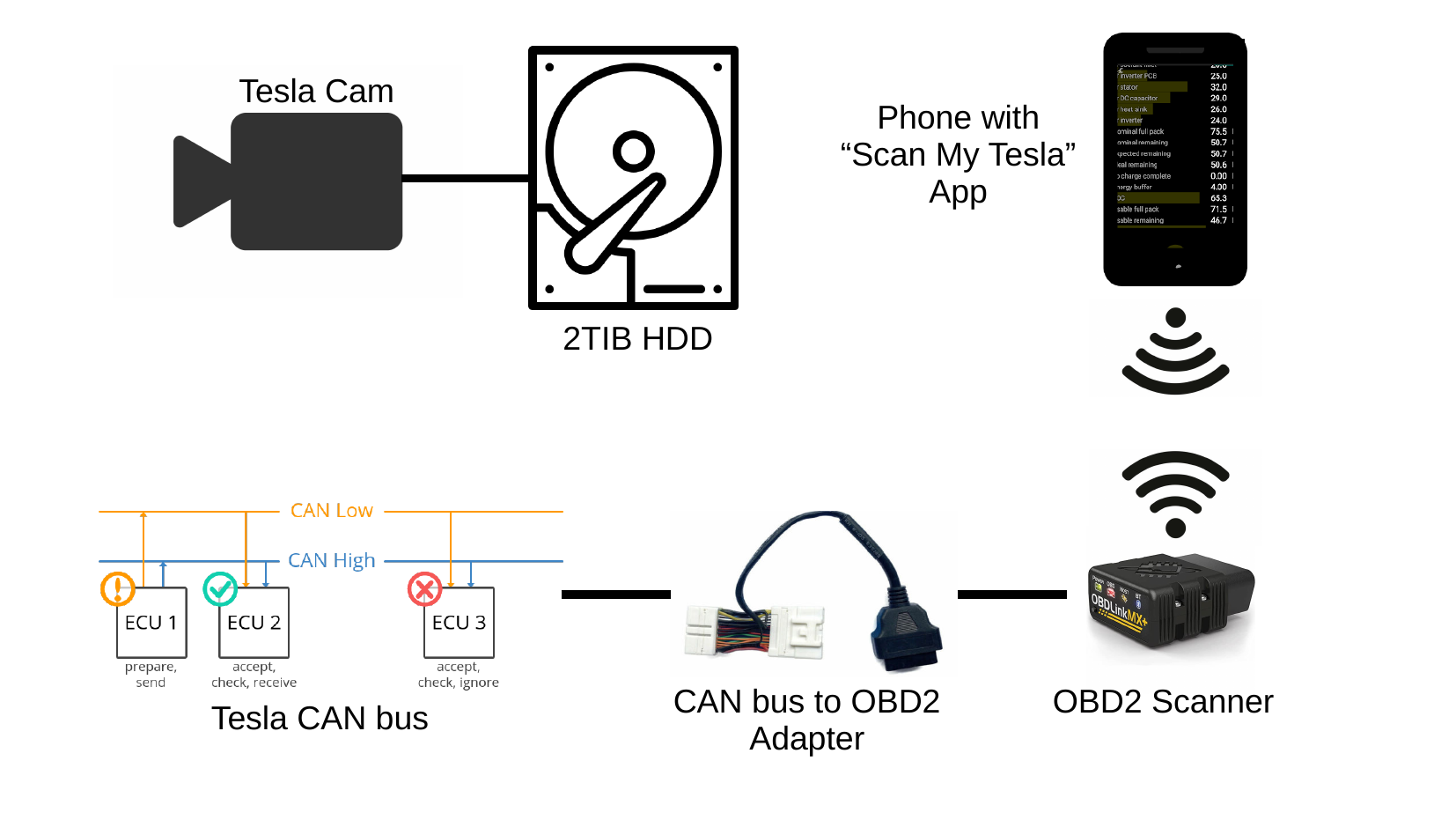}
  \caption{Data Collection Process}
\end{figure}
Initially, the Udacity dataset provided by NVIDIA’s Dave 2 system was considered. However, as it does not host data from the Controller Area Network (CAN) bus, data was collected locally instead. This was important because the CAN bus data provides valuable information about the car's state, such as the engine speed, throttle position, and brake pedal position.

The training data collected was diverse in terms of lighting, incline, and traffic conditions. This was done to ensure that the model would be able to generalize to different driving situations. Most of the data collected was on highways between San Jose and San Francisco, California. This was because these roads are representative of the types of roads that the model would be used on. In addition, other types of roads included in the data were driveways, residential streets, and two-lane roads. This was done to ensure that the model would be able to handle a variety of driving conditions.

As seen in Figure 2, a Tesla Model 3 was used to collect the data. Tesla provides a "dashcam" mode that allows a user to connect a storage device to the Tesla in order to obtain a video feed from the on-board cameras. A 2TB hard drive (HDD) was connected to the Tesla through a USB to HDD cable. To obtain the CAN bus data, an OBD2 adapter was connected to a group of OBD2 cables behind the center console of the car. Then, an OBDLink LX was connected to the adapter. Finally, the "Scan My Tesla" app was downloaded to capture the OBD2 data.

When driving the Tesla for data, the hard drive would record the video data and a phone with the "Scan My Tesla" app collected the CAN Bus data. This is to be expected, as the car was driving on mostly straight roads. The video data was recorded at 36 frames per second and there were four camera angles: center front, center back, right passenger area, and left driver area. This was done to capture a complete view of the car's surroundings. The CANbus data refreshed every one-thousandth of a second and there were 248 parameters.

\subsection{Preprocessing}
Due to the two different data streams, each stream had to be processed and then synchronized to the same time frame. The CAN bus data originally had 248 parameters, but most of these were unnecessary as they were not updated or had minimal value. The CAN bus data was trimmed down to 5 parameters as shown in \autoref{tab:1}.

Each parameter was chosen for a different reason. For the battery parameter, it was concluded that power, voltage, and current of the battery could possibly correlate to the car's speed. The consumption and sleep current were also chosen because they were updated regularly, which could help the model have a richer dataset. The CAN bus data was then compressed by a factor of 25 to match the refresh rate of the video.

For the video data, the right, left, and back camera angles were discarded and only the front camera angle was used, as there was sufficient data from one camera. The front camera data consisted of multiple 1-minute videos, so they were joined together using FFMPEG. Then each frame of the new video was turned into an image.

The video and the CAN bus data were synchronized together with an error of a couple milliseconds by finding a landmark where there was no measured movement. When looking for landmarks in the data, I looked for frames where it was evident there was no movement and then I looked for entries where the speed was zero in CAN bus data and I used that point to sync them together.

An issue with the video feed was encountered, as the video data from the Tesla needed to be saved every 20 minutes. This required clicking on an icon to save, and the resulting save took 203 seconds. During this time, CAN bus data was still being captured, which caused a desynchronization between the CAN bus data and the video data. To combat this, each separate save was grouped into individual groups:

\begin{itemize}
  \item Group 1 is mainly residential street with a bit of highway driving
  \item Group 2 is only highway driving
  \item Group 3 is halfway highway driving and half residential driving
  \item Group 4 is only highway driving
  \item Group 5 is mainly residential street with a bit of highway driving
\end{itemize}

\subsection{Model}

Though models are not the focus of this paper, multiple different models were used in order to provide quantitative evidence on how CAN Bus data diminishes loss.
\begin{figure}[h]
    \centering
  \includegraphics[width=.9\columnwidth]{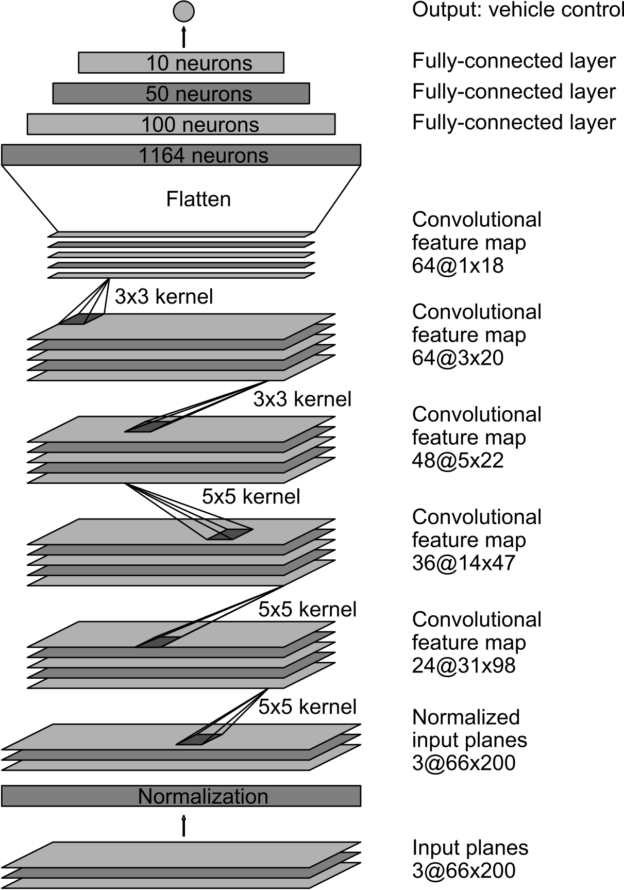}                                              
   \caption{NVIDIA model Visualized}
   \label{fig:nvidia}
 \end{figure}

The first model used was NVIDIA's DAVE-2 model, visualized in \autoref{fig:nvidia}. Though fairly simple when compared to the other models the data was trained on, the NVIDIA model is shown to have lowest error when it comes to end-to-end steering with computer vision. More details about this model can be found in NVIDIA's paper \cite{https://doi.org/10.48550/arxiv.1604.07316}.
 
\begin{figure}[hbt!]
\centering
\includegraphics[width=\columnwidth]{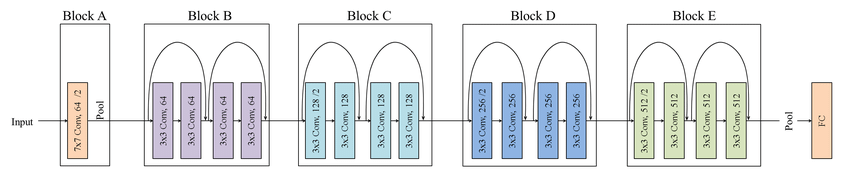}
\caption{RESNET Model Visualized}
\end{figure}
The two RESNET models used were RESNET50 \cite{he2015deep} and RESNET152V2 \cite{DBLP:journals/corr/HeZR016}. The RESNET models were originally trained on images which allows it have lower error when trained on different datasets. Additional benefits include each layer consisting of residual blocks allowing RESNET to keep consistent performance compared to alternate models at the time. In addition, the model included shortcut connections allowing RESNET to keep the parameters from previous layers resulting in much better accuracy and performance. RESENT was groundbreaking as it didn't lose the accuracy with an increasing number of layers. 

More models were also trained on the data. Many other image models were trained on the data due to them being similar to RESNET. This includes NASNETLARGE \cite{DBLP:journals/corr/ZophVSL17}, VGG \cite{simonyan2015deep}, and EfficientNet \cite{DBLP:journals/corr/abs-1905-11946}.

\begin{figure}[h]
  \includegraphics[width=\columnwidth]{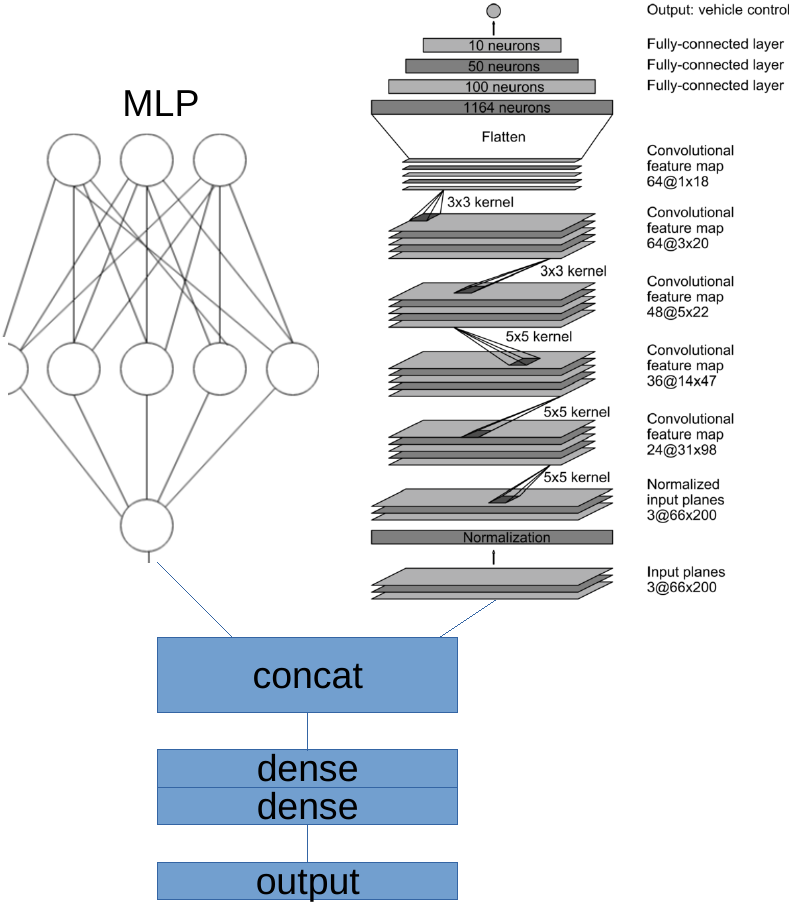}
  \caption{NVIDIA model with CAN bus input}
\end{figure}

\begin{table*}[th!]
  \centering
  \renewcommand{\arraystretch}{1.2}
   
  \begin{tabular}{|c|c|c|c|c|}
    \hline
    & \multicolumn{2}{c|}{\textbf{Without CAN Bus Data}} & \multicolumn{2}{c|}{\textbf{With CAN Bus Data}}\\
    \cline{2-5}
    & \textbf{Training Error} & \textbf{Validation Error} & \textbf{Training Error} & \textbf{Validation Error}\\
    \hline
    NVIDIA&0.01063&0.02492&0.009990&0.01970\\\hline
    RESNET50& 0.03464 & 0.03464 & 0.01230 & 0.01659\\\hline
    EfficientNetB7& 0.04 & 0.04 & 0.01584 & 0.01197  \\ \hline
    VGG19& 0.02299 & 0.01940 & 0.01080 & 0.01477  \\ \hline
    NASNetLarge& 0.01386 & 0.08366 & 0.009690 & 0.01631  \\ \hline
    RESNET152V2& 0.01589 & 0.02035 & 0.006115 & 0.006083  \\ \hline
  \end{tabular}
  \caption{Model RMSE with/without CAN Bus Data}
  \label{tab:2}
\end{table*}
When training the models on the CAN bus data, the data was first piped into a Multi-Layer Perceptron (MLP). The MLP is a type of neural network that is well-suited for classification tasks on numerical data such as the CAN Bus data collected from the car. 

The output of the MLP was then concatenated with the output of the CNN. The concatenated data was then passed through two dense layer and then a final dense layer to produce the output. The output of the final dense layer is a probability distribution over the possible vehicle states.

\section{Results and Discussion}
\subsection{Metrics}

Two different loss functions were used: Mean Square Error (MSE) and Root Mean Square Error (RMSE). RMSE is the root of MSE. MSE was used when training the model, while RMSE was used when comparing the model. Accuracy was not used to compare the models as to determine accuracy the model has to predict the exact Y value. However, since the steering angle was so precise it is difficult to predict to that level of precision. Instead, end to end steering prediction uses root mean square error which determines how far the predicted value is from the actual value. Root Mean square error is hence a more optimal metric for this model.
\begin{equation}
    \text{MSE}=\frac{1}{n}\Sigma(\hat{y}_{i}-y_{i})^2
    \label{eqn:mse}
\end{equation}

\begin{equation}
\text{RMSE}=\sqrt{\frac{1}{n}\Sigma(\hat{y}_{i}-y_{i})^2}
\label{eqn:rmse}
\end{equation}

Also, the adam optimizer was used with a learning rate of 0.0001.

\subsection{Results}

When comparing the results when the models were trained with the CAN Bus data and without it as shown in \autoref{tab:2}, there is a variance in decrease of error. An example of the variance is that RESNET152V2's error dropped by about 70\% for the validation loss while NVIDIA only dropped by 39\% for its validation loss. The reason for the discrepancy for the decrease in error is most likely due to the structure of the model. For example, the NVIDIA model was made to work specifically on only image data for steering angle prediction while the RESNET152V2 model was made to work on a variety of datasets which could have been the reason why it performed so well with the CAN Bus Data. Though it is still shown that overall, supplementing existing computer vision models with CAN Bus data decreases the error. 

The results of the study show that adding CAN bus data to the training data reduces the loss of the model. This means that the model is able to learn more accurately when it is trained with CAN bus data. This is likely because CAN bus data provides more information about the vehicle's behavior than the other data that was used to train the model.

\subsection{Discussion}
During the collection process, many errors were encountered. Initially, multiple pieces of software such as Inpa had to be downloaded in order to interface with the CAN bus. Even though data could be collected from the CAN bus, the steering angle could not be accessed. This led to significant time  trouble shooting which eventually led to discarding the approach of gathering CAN bus data through the ODB2 port as the it was deemed that pin in the OBD2 port was most likely broken. This would lead to some of the data not showing up when extracting the CAN Bus Data. As an alternative, CAN bus data was acquired by man in the middling the CAN Bus wires. An OBD2 adapter was needed to extract the CAN bus data straight from the wires. Finally, setting up the CAN bus extraction process was time consuming; however, benefits outweigh the costs. 

\section{Conclusions and Future Work}
To increase the robustness of the model in real world applications, it is important to collect data from a variety of weather conditions and seasons. This will allow the model to learn how to perform well in different environments, such as sunny, rainy, and snowy days, as well as different seasons. The model will be able to handle different lighting conditions, road surfaces, and other factors that can affect its performance.

In addition to collecting data from a variety of weather conditions and seasons, data augmentation can also improve the performance of the model. Data augmentation is a technique that artificially increases the size of the dataset by creating new data points from existing data. This can be done by applying transformations to the data, such as adding noise, cropping the images, or rotating them. It helps to prevent overfitting by providing the model with more data to learn from which allows the model to generalize to new situations by making it more robust to variations in the data.

The performance of the model can also be improved by breaking down the CAN Bus data into individual components and training each component on the same model. This will allow us to see which components of the CAN Bus data are most important for predicting the steering angle. This information can be used to improve the model's accuracy and make it more robust to changes in the data.

Additionally, the model can be trained unsupervised on the CAN Bus data. This means that model will not be provided with any labels for the steering angle. The model will have to learn to predict the steering angle from the CAN Bus data on its own. This is a more challenging task, but it can lead to a more accurate and robust model.

Unsupervised learning is a valuable technique for self-driving cars because it allows the model to learn from unlabeled data. This is important because there is a lot of unlabeled data available, such as data from sensors that are not used for steering. Unsupervised learning can help the model to learn from this data and improve its performance in new situations. This would be a valuable addition to the model, as it would allow it to learn from a much larger dataset. This would make the model more robust to noise and other factors that can affect its performance.

Finally, other sources of data, such as GPS and online weather data, could be added to the model. This would allow the model to learn from a variety of sources, which would further improve its performance. GPS data can provide information about the car's location, which can be used to improve the model's accuracy in different parts of the world. For example, the model can learn to take into account the incoming turns which would help it predict future steering angles. Online weather data that can provide information about the current weather conditions, which can be used to improve the model's performance in different weather conditions. For example, the model can learn to adjust its steering angle for slippery roads or poor visibility. By adding additional data to the model, the model can learn to take into account more factors when making predictions. This would allow the model to be more accurate in a variety of situations.

Overall, there is still a lot of room for improvement in the future, and these are just a few of the ways that the model could be enhanced. By collecting more data, using different models, and adding more sources of data, the model could be made more robust and accurate.  The more data the model has to learn from, the better it will be able to make predictions. This is why it is important to collect data from a variety of sources, including different weather conditions, seasons, and driving environments. 

\bibliographystyle{IEEEtran}

\begin{thebibliography}{10}
\providecommand{\url}[1]{#1}
\csname url@samestyle\endcsname
\providecommand{\newblock}{\relax}
\providecommand{\bibinfo}[2]{#2}
\providecommand{\BIBentrySTDinterwordspacing}{\spaceskip=0pt\relax}
\providecommand{\BIBentryALTinterwordstretchfactor}{4}
\providecommand{\BIBentryALTinterwordspacing}{\spaceskip=\fontdimen2\font plus
\BIBentryALTinterwordstretchfactor\fontdimen3\font minus \fontdimen4\font\relax}
\providecommand{\BIBforeignlanguage}[2]{{%
\expandafter\ifx\csname l@#1\endcsname\relax
\typeout{** WARNING: IEEEtran.bst: No hyphenation pattern has been}%
\typeout{** loaded for the language `#1'. Using the pattern for}%
\typeout{** the default language instead.}%
\else
\language=\csname l@#1\endcsname
\fi
#2}}
\providecommand{\BIBdecl}{\relax}
\BIBdecl

\bibitem{https://doi.org/10.48550/arxiv.1504.01716}
\BIBentryALTinterwordspacing
B.~Huval, T.~Wang, S.~Tandon, J.~Kiske, W.~Song, J.~Pazhayampallil, M.~Andriluka, P.~Rajpurkar, T.~Migimatsu, R.~Cheng-Yue, F.~Mujica, A.~Coates, and A.~Y. Ng, ``An empirical evaluation of deep learning on highway driving,'' 2015. [Online]. Available: \url{https://arxiv.org/abs/1504.01716}
\BIBentrySTDinterwordspacing

\bibitem{7995975}
Z.~Chen and X.~Huang, ``End-to-end learning for lane keeping of self-driving cars,'' in \emph{2017 IEEE Intelligent Vehicles Symposium (IV)}, 2017, pp. 1856--1860.

\bibitem{https://doi.org/10.48550/arxiv.1604.07316}
\BIBentryALTinterwordspacing
M.~Bojarski, D.~Del~Testa, D.~Dworakowski, B.~Firner, B.~Flepp, P.~Goyal, L.~D. Jackel, M.~Monfort, U.~Muller, J.~Zhang, X.~Zhang, J.~Zhao, and K.~Zieba, ``End to end learning for self-driving cars,'' 2016. [Online]. Available: \url{https://arxiv.org/abs/1604.07316}
\BIBentrySTDinterwordspacing

\bibitem{https://doi.org/10.48550/arxiv.1704.07911}
\BIBentryALTinterwordspacing
M.~Bojarski, P.~Yeres, A.~Choromanska, K.~Choromanski, B.~Firner, L.~Jackel, and U.~Muller, ``Explaining how a deep neural network trained with end-to-end learning steers a car,'' 2017. [Online]. Available: \url{https://arxiv.org/abs/1704.07911}
\BIBentrySTDinterwordspacing

\bibitem{Sallab_2017}
\BIBentryALTinterwordspacing
A.~E. Sallab, M.~Abdou, E.~Perot, and S.~Yogamani, ``Deep reinforcement learning framework for autonomous driving,'' \emph{Electronic Imaging}, vol.~29, no.~19, pp. 70--76, jan 2017. [Online]. Available: \url{https://doi.org/10.2352\%2Fissn.2470-1173.2017.19.avm-023}
\BIBentrySTDinterwordspacing

\bibitem{https://doi.org/10.48550/arxiv.1912.05440}
\BIBentryALTinterwordspacing
S.~Du, H.~Guo, and A.~Simpson, ``Self-driving car steering angle prediction based on image recognition,'' 2019. [Online]. Available: \url{https://arxiv.org/abs/1912.05440}
\BIBentrySTDinterwordspacing

\bibitem{9011341}
N.~Warakagoda, J.~Dirdal, and E.~Faxvaag, ``Fusion of lidar and camera images in end-to-end deep learning for steering an off-road unmanned ground vehicle,'' in \emph{2019 22th International Conference on Information Fusion (FUSION)}, 2019, pp. 1--8.

\bibitem{karkus2022diffstack}
P.~Karkus, B.~Ivanovic, S.~Mannor, and M.~Pavone, ``Diffstack: A differentiable and modular control stack for autonomous vehicles,'' 2022.

\bibitem{he2015deep}
K.~He, X.~Zhang, S.~Ren, and J.~Sun, ``Deep residual learning for image recognition,'' 2015.

\bibitem{DBLP:journals/corr/HeZR016}
\BIBentryALTinterwordspacing
------, ``Identity mappings in deep residual networks,'' \emph{CoRR}, vol. abs/1603.05027, 2016. [Online]. Available: \url{http://arxiv.org/abs/1603.05027}
\BIBentrySTDinterwordspacing

\bibitem{DBLP:journals/corr/ZophVSL17}
\BIBentryALTinterwordspacing
B.~Zoph, V.~Vasudevan, J.~Shlens, and Q.~V. Le, ``Learning transferable architectures for scalable image recognition,'' \emph{CoRR}, vol. abs/1707.07012, 2017. [Online]. Available: \url{http://arxiv.org/abs/1707.07012}
\BIBentrySTDinterwordspacing

\bibitem{simonyan2015deep}
K.~Simonyan and A.~Zisserman, ``Very deep convolutional networks for large-scale image recognition,'' 2015.

\bibitem{DBLP:journals/corr/abs-1905-11946}
\BIBentryALTinterwordspacing
M.~Tan and Q.~V. Le, ``Efficientnet: Rethinking model scaling for convolutional neural networks,'' \emph{CoRR}, vol. abs/1905.11946, 2019. [Online]. Available: \url{http://arxiv.org/abs/1905.11946}
\BIBentrySTDinterwordspacing

\end{thebibliography}

\end{document}